\documentclass[letterpaper, 10 pt, conference]{ieeeconf}  

\IEEEoverridecommandlockouts                              

\usepackage{amsmath} 
\usepackage{amssymb}  
\usepackage{siunitx}
\usepackage[dvipsnames]{xcolor}
\usepackage{esvect}
\usepackage{multirow}
\usepackage{graphicx}
\usepackage{mathtools}
\usepackage{float}
\usepackage[font=small,skip=1pt,belowskip=1pt]{subcaption}
\usepackage[font=small,skip=1pt,belowskip=1pt]{caption}
\usepackage{cite}

\title{\LARGE \bf
RCA: Ride Comfort-Aware Visual Navigation via Self-Supervised Learning
}

\usepackage{graphicx}
\usepackage{hyperref}
\usepackage{keyval}
\usepackage[ruled,vlined]{algorithm2e}
\usepackage{amsmath}
\usepackage{amssymb}
\usepackage{cleveref}
\usepackage{fancyhdr} 
\usepackage{color}
\usepackage{comment}

\newcommand{\idest}{{\it i.e.}, }
\newcommand{\exempli}{{\it e.g.}, }

\fancypagestyle{firststyle}
{
  \fancyhf{}

}

\fancypagestyle{secondstyle}
{
  \fancyhf{}
  \fancyhead[R]{\footnotesize \thepage}

}
\Crefname{equation}{Eq.}{Eqs.}
\Crefname{figure}{Fig.}{Figs.}

\author{Xinjie Yao, Ji Zhang, Jean Oh
\thanks{All authors are with the Robotics Institute at Carnegie Mellon University, Pittsburgh, PA 15213. Emails: {\tt\small \{xinjieya, zhangji, hyaejino\}@andrew.cmu.edu}}%
}

\begin{document}

\maketitle
\thispagestyle{empty}
\pagestyle{empty}

\begin{abstract}

Under shared autonomy, wheelchair users expect vehicles to provide safe and comfortable rides while following users' high-level navigation plans. To find such a path, vehicles negotiate with different terrains and assess their traversal difficulty. Most prior works model surroundings either through geometric representations or semantic classifications, which do not reflect perceived motion intensity and ride comfort in downstream navigation tasks. We propose to model ride comfort explicitly in traversability analysis using proprioceptive sensing. We develop a self-supervised learning framework to predict traversability costmap from first-person-view images by leveraging vehicle states as training signals. Our approach estimates how the vehicle would ``feel'' if traversing over based on terrain appearances. We then show our navigation system provides human-preferred ride comfort through robot experiments together with a human evaluation study.
\end{abstract}

\section{INTRODUCTION}

There are more than 2.7 million people in the United States who primarily use wheelchairs for their mobility on a daily basis, and the number of wheelchair users is also expected to grow\cite{wheelchair_stats}. With the advancement of the autonomous vehicle technology, autonomous wheelchairs are gradually becoming attainable~\cite{leaman_2017_trhms}. Previous studies show that having the sense of control is important to human users, \idest they prefer systems that provide high-level controls over fully autonomous ones, promoting the concept of shared autonomy~\cite{muelling2015autonomy,selvaggio2021autonomy}. In this context, our work addresses low-level robotic vehicle control given human users' high-level control inputs. Specifically, we focus on finding safe and comfortable paths that generally follow a user's high-level navigation plan. 

Finding a collision-free, smooth path has been a central challenge in robot navigation. 
3D-based approaches, as in~\cite{chao2021}, interpret a surrounding environment as rigid 3D objects and estimate the traversabilitiy based on their geometric properties. While these approaches can detect obstacles and the general traversability of terrain well for large-sized vehicles with robust mobility, they lack semantic understanding of social norms as well as subtle differences in traversability that are important for wheelchair applications.

The other stream of prior works relies on semantic classification of environments, associating each predefined semantic class with a traversal difficulty. Existing works on this end require fine-grained terrain semantic classes and merge them into levels of traversability\cite{Viswanath2021OFFSEGAS}. In these approaches, the entire region with the same class label is assigned the same traversability cost, ignoring the variations within each region. Moreover, such a terrain analysis depends heavily on human assessment of traversing difficulty which should be customized for various purposes of navigation, including wheelchair applications. 

To address these shortcomings of existing approaches, we propose a first-person image-based approach where traversability is defined as riding comfort and can be measured in terms of the vehicle state. As in the semantic classification methods, the main idea is to use vision to estimate the traversability. Instead of requiring human-defined semantic labels, our framework learns to associate the visual input with how the vehicle state changes when the robot is actually traversing over the spot. Because our approach's learning signal comes directly from the vehicle state, we propose a self-supervised learning approach for learning this mapping between vision and internal vehicle state and, eventually, traversability cost. To enable a direct mapping from vision to cost, we use unsupervised learning as in~\cite{Zurn2021SelfSupervisedVT} to first classify vehicle states into coarse categories. Whereas \cite{Zurn2021SelfSupervisedVT} uses this idea for semantic classification and image segmentation, we propose a method for generating a continuous traversability cost given an image input.  


We fully integrate the proposed approach (RCA) on a robotic wheelchair platform for our experiments and evaluation. Based on human evaluation, the proposed approach is strongly preferred to the 3D-based and semantic classification methods according to stability, path normality, safety, trustworthiness, and preference. 

Our contributions are summarized as follows:
\vspace{-2pt}
\begin{enumerate}
    \item we propose to define ride comfort in terms of proprioceptive sensing and include that in traversability analysis to improve the general quality of navigation paths under shared autonomy; 
    \item we develop a self-supervised learning framework (RCA) for predicting a traversability costmap based on first-person-view image inputs; 
    \item we show that human assessment agrees with the normalized Perceived Vehicle Motion Intensity (PMI) score, supporting the intuition that proprioceptive data can capture the traversability and ride comfort; and
    \item we share our findings from a human evaluation study on how humans determine the ride quality according to various aspects including vehicle dynamics and perceived motion intensity.
\end{enumerate}
\vspace{-5pt}

\section{RELATED WORK}

Conventional approaches of vehicle navigation model the environments through geometric expressions and obtain outstanding results with rigid objects. Representations of the environments involve 3D pointcloud~\cite{chao2021}, signed distance field~\cite{oleynikova2016signed}, and 2.5D elevation map\cite{ChavezGarcia2018LearningGT}. They do not take into account the complexity of vehicle-terrain interactions in which deformable and nonrigid terrains bring additional ambiguity into the measurement of traversability. 

With the advancement of the urban-scene image segmentation task\cite{cityscape_2016_CVPR}, semantic-based approaches demonstrate promising results due to the rich texture information from vision sensors. Traversal difficulty is typically modeled by the visual appearance of terrains, namely via either mapping semantic terrain classes to traversability \cite{Viswanath2021OFFSEGAS} \cite{Otsu2016AutonomousTC} \cite{DBLP_Gao} or defining a texture-dependent score function to reflect difficulties \cite{Kim2006TraversabilityCU} \cite{safe_simone_2020}. The association between visual distinctness and traversability is constrained to the vision domain and is commonly assessed by human experts, whose subjective interpretations could be erratic given the complex dynamics. 

Another stream of emerging works deploy end-to-end learning systems through Inverse Reinforcement Learning\cite{Zhu2020OffroadAV}, Reinforcement Learning applied to model predictive controller\cite{quad_locomotion}, and Imitation Learning from experts using additional sensory inputs\cite{DBLP:conf/rss/PanCSLYTB18}. These approaches generally require a large amount of vehicle driving data in real world or in high fidelity simulators and may suffer from the vehicle hazards caused by the aggressive exploration behaviors\cite{DBLP:BAGR}.

Apart from leveraging the single modality of the sensory inputs, self-supervised learning from cross-modality sensor inputs shows successful practices to supplement the complementary information across modalities\cite{9043710}. It investigates the learning of associations between one modality and labels generated from other modalities. During deployment, the predictive model estimates the associate labels at the absence of other modalities. Proprioceptive data are typically generated from traversing actual terrains and are linked to its exteroceptive data for future prediction under this reciprocal manner. Approaches using force-torque signals \cite{Wellhausen2019WhereSI}, vehicle vibrations \cite{Brooks2012SelfsupervisedTC}\cite{8170440}, and acoustic signals \cite{Zurn2021SelfSupervisedVT}\cite{kris2021recog}\cite{DBLP:journals/corr/abs-1804-00736} have been proposed to construct this cross-modal association with terrain visual inputs.  
Vehicle dynamics based approaches fail to leverage the texture information provided from vision images to create labels, and instead only translate proprioceptive signals to the vision domain\cite{Wellhausen2019WhereSI}\cite{Brooks2012SelfsupervisedTC}\cite{8170440}. Approaches using acoustic waveform \cite{Zurn2021SelfSupervisedVT}\cite{kris2021recog}\cite{DBLP:journals/corr/abs-1804-00736} regard terrain properties as semantic classes and formulate the navigation task as semantic classification\cite{Viswanath2021OFFSEGAS}. They fail to illustrate vehicle dynamics in the feature space and are categorized as a semantic-based approach. One recent approach \cite{Zurn2021SelfSupervisedVT} selects acoustic signal triplets based on the Euclidean distance between visual embeddings. With a Siamese encoder, it obtains a distinct feature space using the acoustic signal reconstruction loss and triplet loss. Those features are further clustered into semantic classes for downstream segmentation task. 

While prior works mostly represent terrain traversability with semantic labels or geometry properties, the proposed RCA approach includes ride comfort in assessing traversal cost and determines vehicle dynamics and perceived motion intensity by complimenting information from the exteroceptive and propriceptive sensing via self-supervised learning.

\section{PROBLEM FORMULATION}\label{sec:prob}
The target problem is predicting the navigation costs using first-person-view monocular camera images. 
Let $\mathcal{i}_t \in \mathcal{I}$ denote an image input at time step $t$ where $\mathcal{I}$ denotes the projection space of the monocular camera images. We aim to find a function that maps this projection space to a 2D cost map, denoted by $\Phi(\mathcal{I}) \rightarrow \mathbb{R}^2$.

Based on the intuition that traversability directly affects the vehicle state, we formulate this problem as finding two mapping functions. Here, we introduce the vehicle state as an internal representation of the cost as follows.  
Let $\mathbf{s}_{\text {t}}$ denote the vehicle state at time $t$, consisting of the 3D robot pose including position, orientation, angular velocity, and linear acceleration, denoted by $\mathbf{s}_{\text {t}}=\left[\mathbf{p}_{\text {t}} \in  \mathbb{R}^3, \mathbf{q}_{\text {t}} \in \mathbf{S O}(\mathbf{3}),  \mathbf{\dot{q}}_{\text {t}},  \mathbf{\ddot{p}}_{\text {t}}\right]$, respectively. 
The first subproblem is to find a mapping function $\Phi_{I \rightarrow S}(\mathcal{I}) \rightarrow \mathcal{S}$ that can predict the vehicle states from input images, minimizing the error between the predicted and the true states.
The second mapping function,
$\Phi_{S \rightarrow \mathbb{R}}(\mathcal{S}) \rightarrow \mathbb{R}^2$, maps the vehicle states to traversability costs, quantifying the traversability based on vehicle states. 
Using this formulation, our self-supervised approach automatically generates training data in the form of an image and the corresponding traversability costmap. This training data can then be used to learn the mapping function from image to cost $\Phi(\mathcal{I}) \rightarrow \mathbb{R}^2$. 

\section{APPROACH}

We first describe the overall navigation system and an overview of the proposed approach in~\Cref{subsec:system} and~\Cref{subsec:overview}, followed by technical details on key modules of the proposed approach.  

    \begin{figure}[htpb]%
    \centering%
    \vspace{-5pt}
      \includegraphics[width=0.9\linewidth]{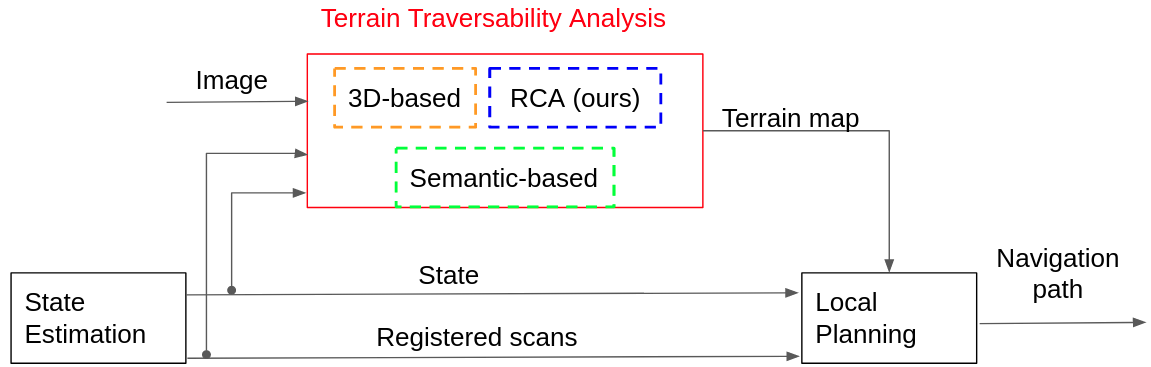}%
    \caption{A navigation system with the terrain analysis module.}%
    \vspace{-15pt}%
    \label{fig:system}%
    \end{figure}

    \begin{figure*}[t!]%
    \centering%
      \includegraphics[width=0.8\linewidth]{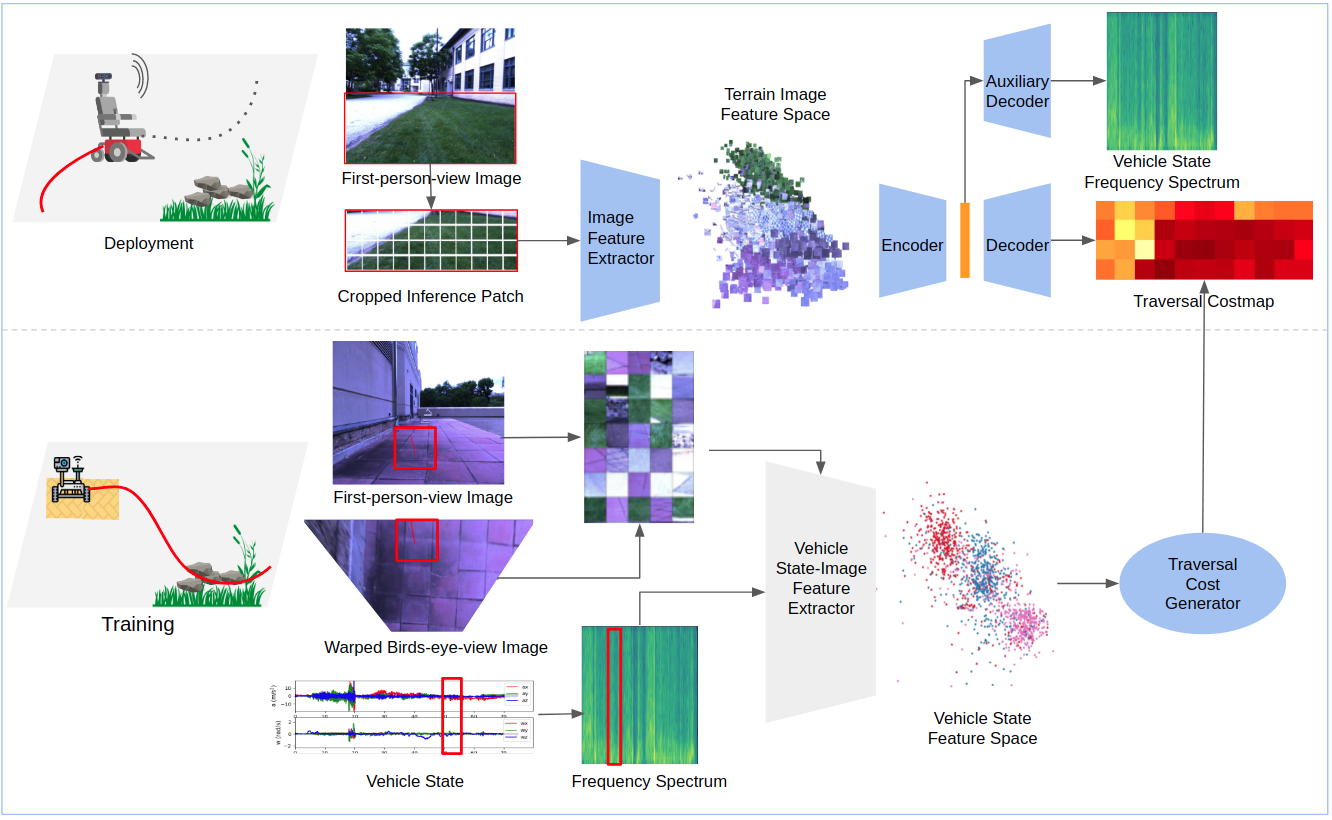}%
      \caption{An overview of RCA. During training, the vehicle experiences physical vibrations by traversing various types of terrains. Using recorded vehicle states and first-person view images, the system learns to estimate traversability cost from images. 
      Instead of manual labeling, we first use unsupervised learning 
      to coarsely group vehicle states into clusters based on vehicle-terrain interactions. Next, with those clusters, we define a continuous cost function to reflect the traversability based on vehicle dynamics. Finally, we train a prediction model to associate terrain images directly to the traversability costs. During deployment, the learned model estimates a terrain costmap to support autonomous navigation.}%
    \vspace{-15pt}%
      \label{fig:framework}%
    \end{figure*}
\subsection{A Navigation System} \label{subsec:system}

    \Cref{fig:system} shows an overall navigation system where the proposed approach contributes to the terrain traversability analysis module.

\vspace{0.1cm}
\noindent\textbf{The State Estimation Module} consists of a data pipeline\cite{loam} that processes 3D lidar scans, images, and inertial measurements. It generates vehicle motions and registers laser scans based on the estimated pose. We note that only the vehicle state from this module constitutes the vehicle state used in our approach as defined in \Cref{sec:prob} and the rest is included for supporting the baselines.
\vspace{0.1cm}

\noindent\textbf{The Terrain Traversability Analysis Module} 
    takes data from onboard sensors, \exempli camera or 3D sensor, to produce a navigation costmap. Our RCA approach is specifically designed for a monocular camera only setting, but we include a 3D-based approach~\cite{chao2021} as well as a semantic classifier approach~\cite{deeplabv3plus2018} as the baselines for evaluation. We also assume that standard vehicle state information such as 6-DOF pose of the vehicle is available. The Terrain Traversability Analysis Module outputs a local terrain map indicating how traversable local region is either in a discrete or continuous fashion.
\vspace{0.1cm}

\noindent \textbf{The Local Planning Module} takes charge of avoiding obstacles in the vicinity of the vehicle, providing low-level planning. This module 3D projects first-person view image-based terrain traversability analysis output onto the registered lidar scans and uses it for collision avoidance planning. The planning algorithm generates collision-free paths based on a trajectory library\cite{localplanner}.



\subsection{An Overview of RCA Approach}\label{subsec:overview}
\Cref{fig:framework} describes RCA framework for predicting a traversability costmap using a monocular camera only. 
Here, we propose a self-supervised learning approach for the terrain traversability analysis module by exploiting a robot's vehicle state as an additional input to the learning algorithm in addition to the image input. Intuitively, our approach aims to learn to predict how a robot would ``feel'' based on the first-person view of the terrain ahead of the robot. We then translate the vehicle state to a continuous numerical cost function using self-supervised learning. By using proprioceptive data as the learning signal for traversability, the model learns to map visual inputs to traversability costs. 

\subsection{Input Data Preprocessing} \label{subsec:data_preprop}
Our approach takes the following two types of inputs: the vehicle states and the first-person view camera images. This pair of data can be collected in an unsupervised manner, \exempli by having a robot explore an environment. Such pairs, however, generally result in a data imbalance issue between these two types of data. More specifically, perspective projection distorts further regions where vehicle-terrain interactions happen and thus are discarded. To address this issue we augment the training data. 

    \subsubsection{Vehicle State Data}
    To model experienced disturbances from vehicle-terrain interaction, we take the angular velocity along the roll and pitch axes $\mathbf{\dot{q}^r}, \mathbf{\dot{q}^p}$, linear acceleration along the Z axis (pointing downward) $\mathbf{\ddot{p}^z}$ as inputs. Signals from three axes are truncated into a window of $n$ samples, and their frequency components are analyzed by the Fast Fourier Transform (FFT) to obtain amplitude spectra $\mathbf{A}^d$ respectively. Finally, three amplitude spectra are concatenated into a single vector.
    
    \subsubsection{Image Data}
    As the vehicle traverses only sub-regions captured by the image, it is essential to obtain the image of those sub-regions. Thus, we project the vehicle trajectory into the distortion-corrected image, and keep footprints with the similar heading direction as the viewpoint. To account for scale ambiguity caused by perspective projection, we keep the closest $m$ footprints. Around each footprint, we crop a box sizing $w \times h$ to indicate areas traversed by the robot and associate them with corresponding vehicle states. For vehicle's footprint $i$, denote $\psi_i$ as the yaw angle, $\mathbf{p^w_i} \in \mathbb{R}^3$ as the position of footprint in the world frame, and $\mathbf{p^c_i}  \in \mathbb{R}^2 $ as the projected footprint in the image frame. Let $\mathbf{K}$ be the camera intrinsic matrix, $\mathbf{R_{w}^{c}} \in \mathbf{S O}(3)$ be the rotation matrix, and $\mathbf{t_{w}^{c}} \in \mathbb{R}^{3}$ be the translation vector from the world frame to the camera frame. The projected vehicle footprints are,
    \begin{equation*}
      \mathbf{p^c_i} = \mathbf{K}[\mathbf{R_{w}^{c}}|\mathbf{t_{w}^{c}}]\mathbf{p^w_i} \quad \forall i \in \{j: |\psi_j-\psi_0| \leq \theta\}
    \end{equation*}
    
    \subsubsection{Data Augmentation}\label{sec:data-augmentation}
    To remedy the data imbalance issue caused by discarding a subset of footprint image patches, we propose to augment the data by generating imaginary images for flat surfaces. We first generate a birds-eye-view (BE-view) image from the first-person-view (FP-view) input, and then take the cropped samples from the BE-view images. The imaginary BE-view images are warped using the Euclidean homography $\mathbf{H_{bc}}$ from the FP-view frame $c$ to the BE-view frame $b$ as follows. For any point in the FP-view image, $\mathbf{p^c}$, the corresponding point in the BE-view image is,
    \begin{equation}
      \mathbf{p^{b}}=\lambda \mathbf{K} \cdot \mathbf{H_{bc}} \cdot \mathbf{K}^{-1} \cdot \mathbf{p^c}
      \label{eq:bev_warp}
    \end{equation}
    where $\lambda$ is a scale factor. Denote $\mathbf{n}$ as the normal of the ground plane, and d as the distance from origin to the plane. We can write the Euclidean homography $\mathbf{H_{bc}}$ as,
    \begin{equation*}
    \mathbf{H_{bc}}=\mathbf{R_{b}^{c}}-\frac{\mathbf{t_{b}^{c}} \mathbf{n}^{T}}{d}
    \end{equation*}
    Corresponding projected footprints are warped using \Cref{eq:bev_warp}. Then image patches for data augmentation are cropped from BE-view images around these transformed footprints.
    
\subsection{Estimating Traversability Cost from Vehicle States} \label{subsec:training}
Our RCA framework only utilizes vehicle states during training as intermediate learning signals. These learning signals are eventually turned into numerical costs. This way, the main learning algorithm uses these predicted costs, along with the images, as self-supervision, rather than utilizing vehicle states directly. This section describes how we estimate traversability costs from vehicle states for that training. The idea is composed of two parts: first, we use an unsupervised learning method as in~\cite{Zurn2021SelfSupervisedVT} to generate coarse labels for vehicle state clusters; second, we use the learned clusters to estimate a continuous cost function given a vehicle state input.  

\subsubsection{Unsupervised Learning for Classifying Vehicle States}\label{subsec:unsupervised}

After input data preprocessing, we have image patches indicating traversal regions and frequency spectra describing experienced motions. As opposed to previous self-supervised vibration-based approaches which handcraft features\cite{Brooks2012SelfsupervisedTC}\cite{8170440}, we exploit the complementary information shared by the visual and vehicle state domains to extract latent features. 

To learn this association, we generally follow the unsupervised feature learning framework proposed in\cite{Zurn2021SelfSupervisedVT}.  
%
%
In essence, this framework brings visually-similar samples closer and visually-distinct samples further away in the target space, \exempli acoustic data in their work.
%
We customize the framework for our interpretation on riding comfort as follows.
First, we extract image features using the Deep Encoding Pooling (DEP) network~\cite{DEP}, which is specialized for ground terrain recognition.
Next, in order to ensure that negative samples are selected from different ground truth classes, we leverage semantic classes clustered in the visual feature space. Such prior knowledge serves as a reference for computing Euclidean distance and selecting negative sample.
%
%
Based on the PCA projection of the feature space, we perform k-Means clustering to obtain coarse labels for terrain classes. 

    \subsubsection{Traversal Cost Generation}\label{cost_generation}
    
    Wheelchair passengers are exposed to greater physiological risks\cite{Gaal1997WheelchairRI} and psychological barriers\cite{comfort_study} with drastic changes in the vehicle motions. Thus we hypothesize that larger and more frequent movements along three dimensions should be avoided, and assigned with higher cost. The amplitude spectra show amounts of motion variations at different frequencies, and vehicle state clusters obtained in \Cref{subsec:unsupervised} serve as a prior to show distinct vehicle dynamics. 
    
    The traversal cost function is considered as a weighted average of roll, pitch, and Z dimension measuring similarity to the average amplitude spectrum of its vehicle state class. Different vehicle state classes are also magnified with a hyperparameter to offset from each other. Let $\mathbf{A^{d,k}_{i}} \in \mathbb{R}^m$ denote the amplitude spectrum along dimension $d$ from the vehicle state class $k$ for the sample $i$. $N^{k}$ describes the number of samples within the vehicle state class $k$. 
    A traversal cost for sample $i$ from vehicle state class $k$ is, 
    \begin{equation}
        T^{k}_{i} = \omega_k \cdot \sum_{d=1}^{3}\mathbf{M^{d,k}} \cdot \mathbf{A^{d,k}_{i}}
    \label{eq:cost}
    \end{equation}
    where $\omega_k$ is a weight parameter of vehicle state class $k$, and where $ \mathbf{M^{d,k}}$ is the mean amplitude spectrum along dimension $d$ from the vehicle state class $k$,
    \begin{equation*}
        \mathbf{M^{d,k}} = \frac{1}{N^{k}}\sum_{i=1}^{N^{k}} \mathbf{A^{d,k}_{i}}
    \end{equation*}
    
    \subsection{Self-supervised Traversal Cost Prediction} \label{subsec:self-supervised}
    
    With the traversal cost function, we associate terrain visual appearances with motion measurements. We implement a encoder-decoder network, to estimate amplitude spectra of vehicle states and to predict traversal costs labeled by \Cref{eq:cost}. 

    To align the visual feature space with the vehicle state feature space, terrain patches are first extracted by the same feature extractor in \Cref{subsec:unsupervised}. Then terrain visual features are further encoded. During training, two decoders start back propagation at different epochs. In early epochs, the encoded space is translated toward vehicle states as only weights from the auxiliary vehicle state decoder get updated. Then later, the traversal cost decoder joins and learns from the partially translated feature space.
    We use $\mathcal{L}_{2}$ loss and smooth-$\mathcal{L}_{1}$ loss for regressing amplitude spectrum and traversal cost, respectively.
    \begin{equation*}
        \mathcal{L}_{2}=\left\|g\left(f\left(\mathbf{I}_{i}\right)\right)-\mathbf{A}_{i}\right\|^{2}
    \end{equation*}
    Define $x=h(f(\mathbf{I}_{i}))-T_i$ as the difference between the traversal cost estimation and the computed traversal cost,
    \begin{equation*}
    {smooth}_{\mathcal{L}_{1}}= \begin{cases}0.5x^{2} & \text { if }|x|<1 \\ |x|-0.5 & \text { otherwise }\end{cases}
    \end{equation*}
    The regression loss is written as,
    \begin{equation*}
    \mathcal{L} = \beta \mathcal{L}_{2} + (1-\beta){smooth}_{\mathcal{L}_{1}}
    \end{equation*}

    Given a camera image, it infers the traversal cost for each cropped patch and aggregates them as 2D costmaps.

\section{EXPERIMENTS}
     We first describe the training dataset and implementation details, followed by descriptions of the main robot experiments.

    \begin{figure}[h!]%
    \centering%
    \begin{subfigure}{0.2\textwidth}%
      \centering%
      \includegraphics[height=2.5cm,width=\linewidth]{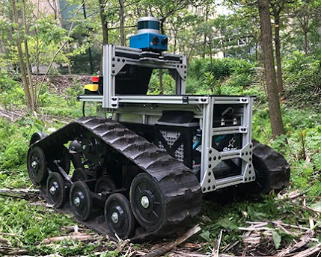}%
      \subcaption{}%
      \label{fig:tracked}%
    \end{subfigure}\;
    \begin{subfigure}{0.2\textwidth}
        \centering%
        \includegraphics[height=2.5cm,width=\linewidth]{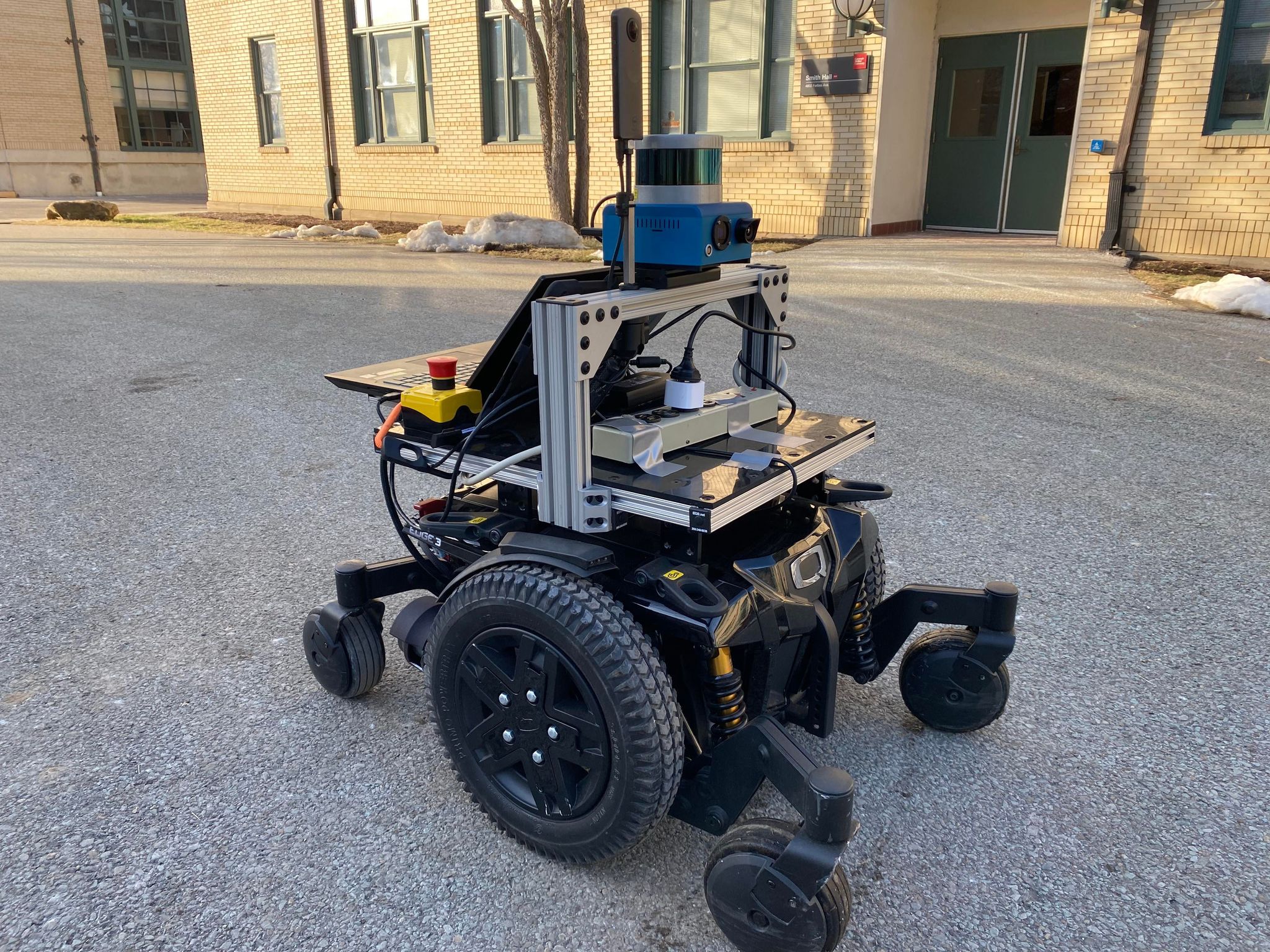}%
        \subcaption{}%
        \label{fig:wheelchair}%
    \end{subfigure}%
    \caption{
    (a) The tracked vehicle used for data collection.
    (b) The robotic wheelchair platform used in the experiments equipped with a 4.1GHz i7 computer and a NVIDIA GTX 1660Ti GPU card. Both platforms are equipped with a Velodyne Puck Lidar, a MEMS-based IMU, and a monocular camera.}%
    \vspace{-20pt}
    \label{fig:vehicle}
    \end{figure}

\subsection{Training the Learning Modules}

    \subsubsection{Dataset} \label{subsec:ds}
    We collected a training dataset in an urban area using a tracked vehicle (\Cref{fig:tracked}) that competently covered a variety of challenging terrains without posing risks to human operators.
    The camera captures frontal images at \SI{5}{\hertz} with 1280 $\times$ 1024 resolution. The state estimation module provides motion measurements at \SI{200}{\hertz}. Our dataset includes 2 hours of operated driving at $\SI{1}{\meter\per\second}$ including variations of soft and hard surfaces with different elevations to ensure a broad spectrum of vehicle motion profiles.

\subsubsection{Implementations} \label{subsec:imple}

    \textit{Estimating Traversability Cost from Vehicle States:} The vehicle state is split into windows of 256 samples and a stride size of 128 samples. The size of the hidden space vector is 32. The image is cropped into patches of 256$\times$256. We adopt Adam Optimizer with a learning rate as $0.001$, a weight decay as $0.0001$, and $\alpha=0.4, \gamma=20$. We trained for 120 epochs using a batch size of 64.
    
     \textit{Self-supervised Traversal Cost Prediction:} The lower half of the input image is cropped into small patches of 256$\times$256 pixels. We adopt Adam Optimizer with a learning rate as $0.0001$ and a weight decay as $0.00001$, training in 150 epochs and batches of 64 each. We set $\beta=1$ during the first 40 epochs and adjust to $\beta=0.4$ for the remaining. 
     As for the k-MEANS clustering, $k=3$ was chosen empirically.
     
\subsection{Robot Experiments}\label{subsec:exp}
To evaluate the quality of RCA, we compare the performance with two baselines: a 3D-based approach~\cite{chao2021} and a semantic classification-based one~\cite{wu2021riss}. To assess the performance of three terrain analysis approaches in actual navigation tasks, we conduct robot experiments with a wheelchair-based vehicle (\Cref{fig:wheelchair}) navigating various terrain conditions. To investigate the performance according to various factors determining ride comfort and to evaluate perceived motion profiles, we design and conduct a human evaluation study.
     
\subsubsection{Experiment Design}
    For evaluation, we record a set of robot videos in a controlled setting such that, 
    for each scenario, the wheelchair robot repeats the same task using the proposed RCA and the two baseline approaches, \idest starting at the same pose towards a short-term goal point specified by a human controller. Each run is repeated three times. The scenarios consists of commonly seen objects and terrains in the urban area.
    
\subsubsection{Perceived Vehicle Motion Intensity (PMI)} 
The PMI Score~\cite{de_winkel_pmi_2020} provides a quantitative measurement on the perception of inertial of motions. It hypothesizes the perception of motions is affected by acceleration, jerk, and their interaction, following a normal distribution with mean as,
    \begin{equation}
    \mu_{\psi}=\omega_{A} A_{\max }+\omega_{J} J_{\max }+\omega_{A J}\left(A_{\max } \times J_{\max }\right)\label{eq:pmi}
    \end{equation}
    where $A_{\max },J_{\max }$ are the maximum of acceleration and jerk for a given motion respectively, and where $\omega_{A}, \omega_{J}, \text{ and } \omega_{AJ}$ are weights on acceleration, jerk and their interactions.
    

\begin{figure*}[tb!]
    \centering
    \raisebox{-0.5\height}{\includegraphics[height=1.9cm, width=.19\linewidth]{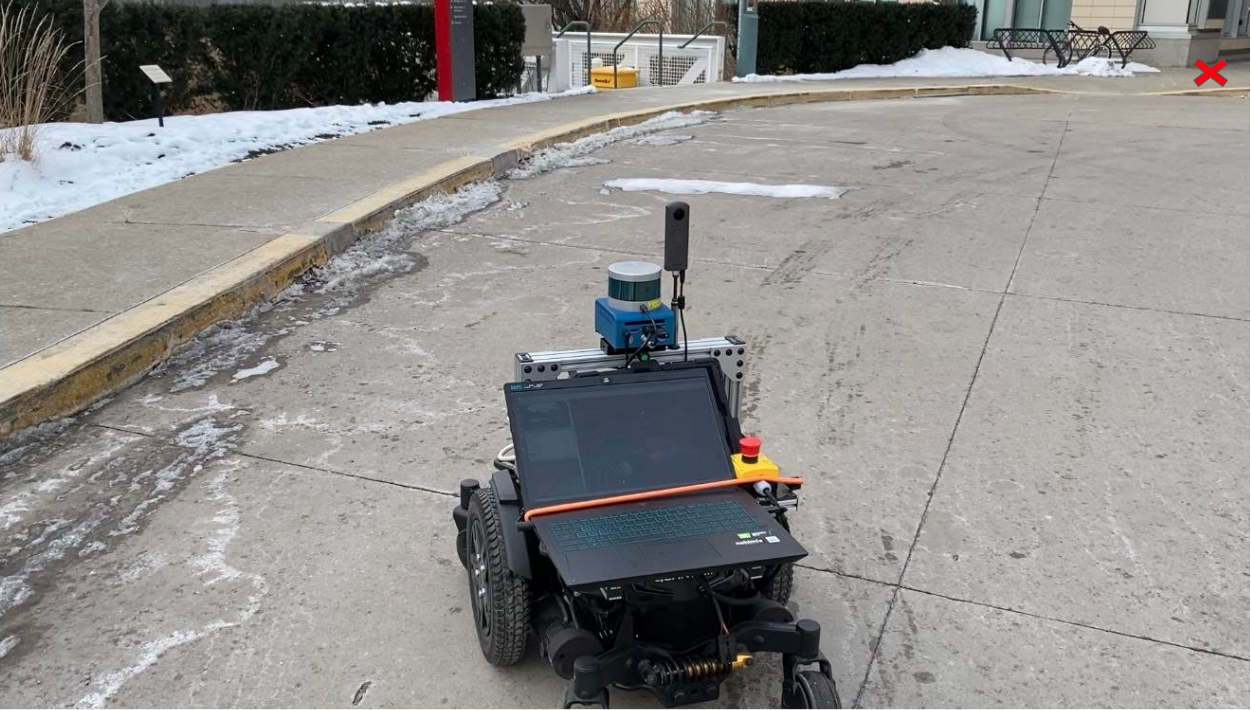}}\;\raisebox{-0.5\height}{\includegraphics[height=1.9cm,width=.19\linewidth]{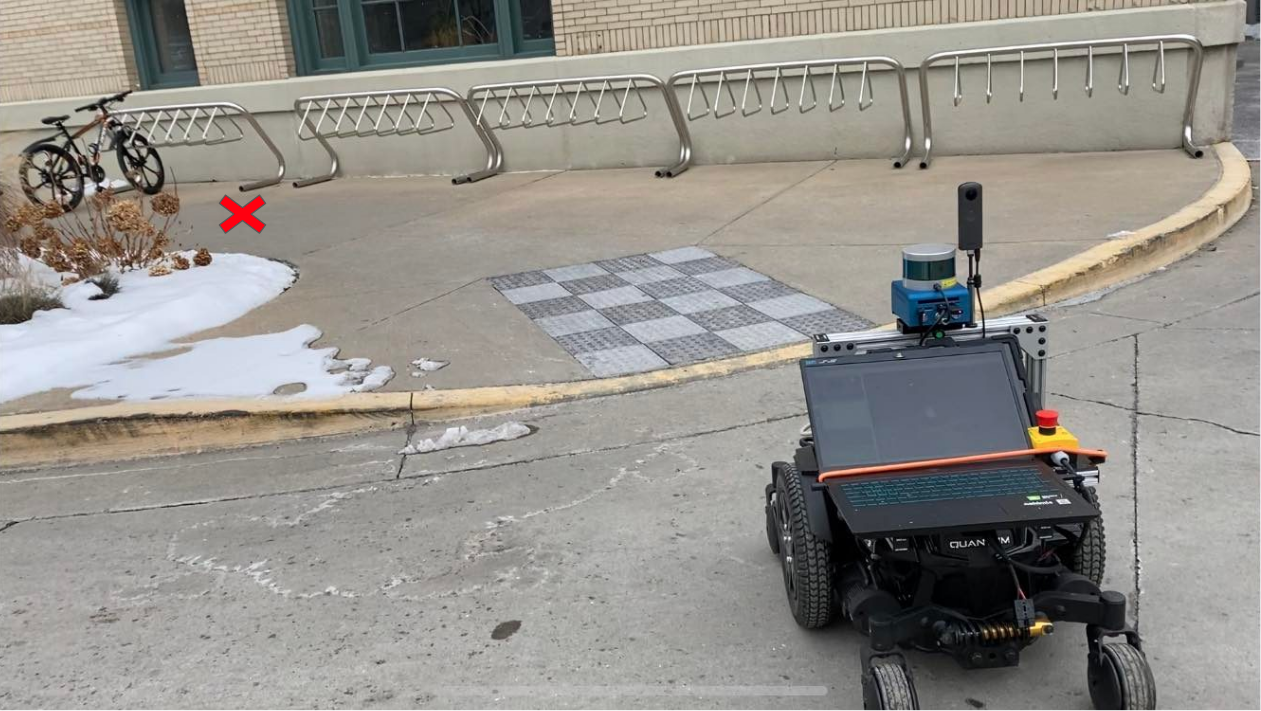}}\;\raisebox{-0.5\height}{\includegraphics[height=1.9cm,width=.19\linewidth]{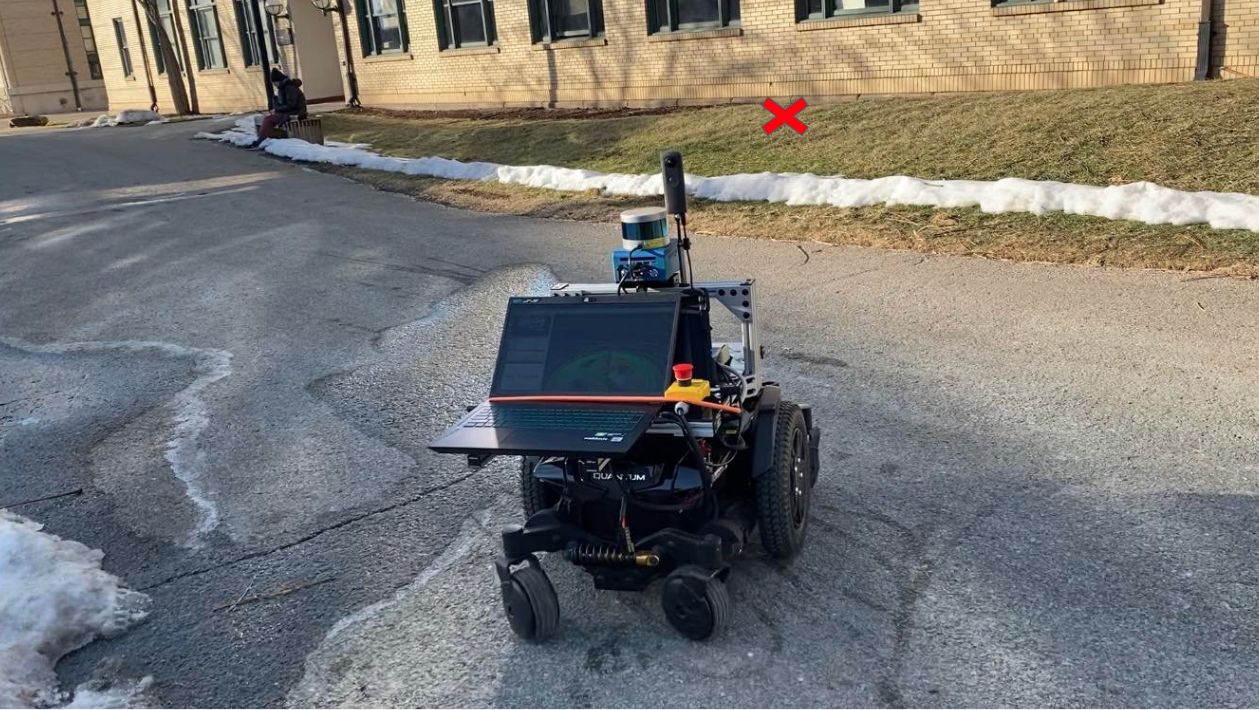}}\;\raisebox{-0.5\height}{\includegraphics[height=1.9cm,width=.19\linewidth]{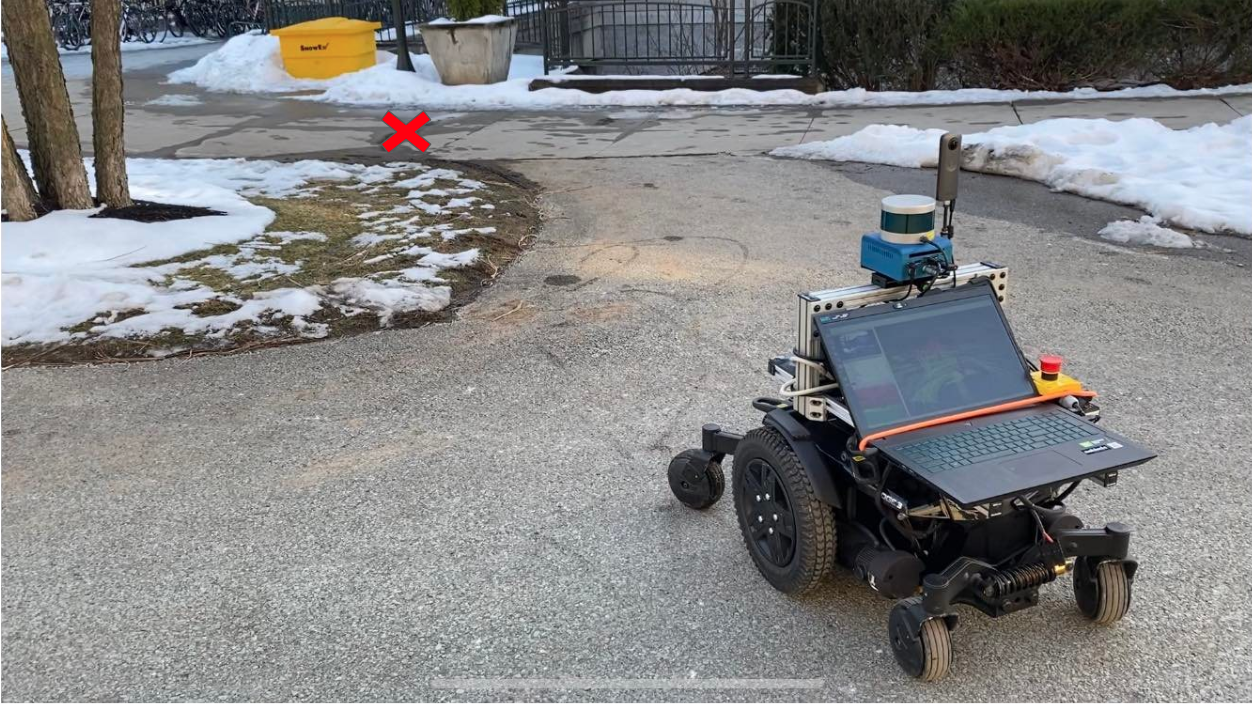}}\;\raisebox{-0.5\height}{\includegraphics[height=1.9cm,width=.19\linewidth]{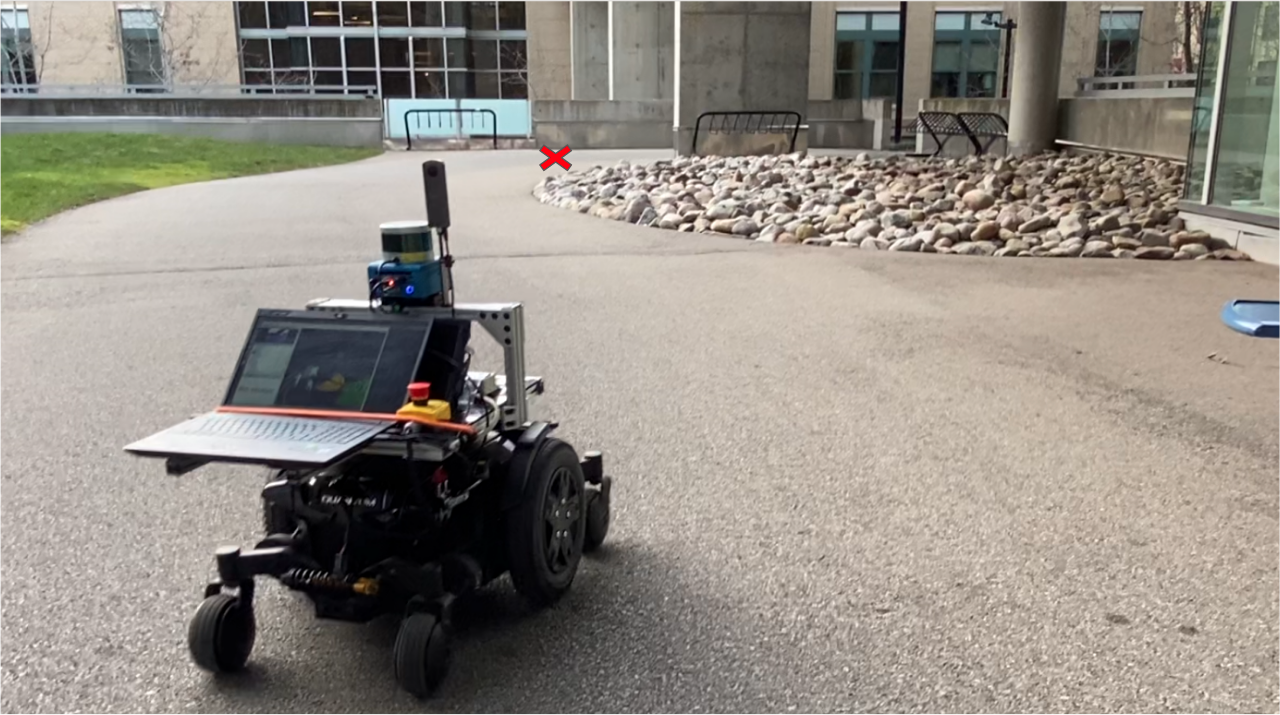}}
       \\ \vspace{.3\baselineskip}
    \raisebox{-0.5\height}{\includegraphics[height=1.9cm,width=.19\linewidth]{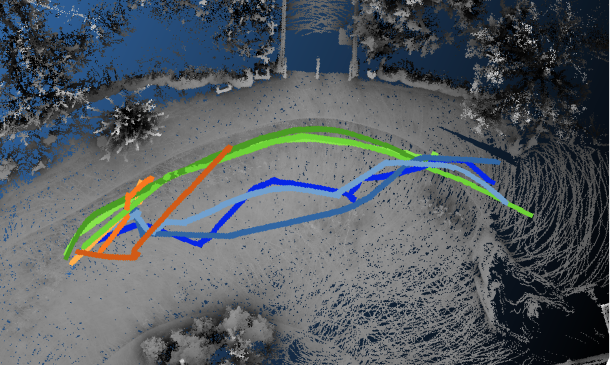}}\;\raisebox{-0.5\height}{\includegraphics[height=1.9cm,width=.19\linewidth]{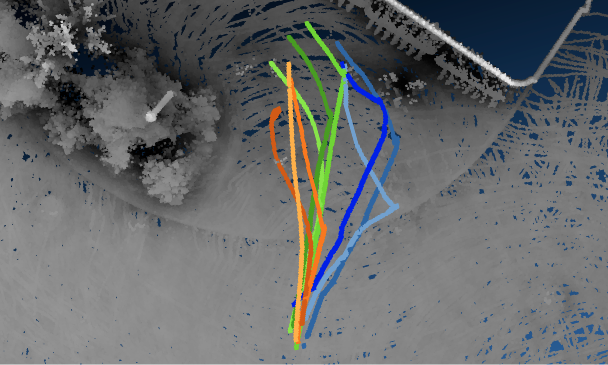}}\;\raisebox{-0.5\height}{\includegraphics[height=1.9cm,width=.19\linewidth]{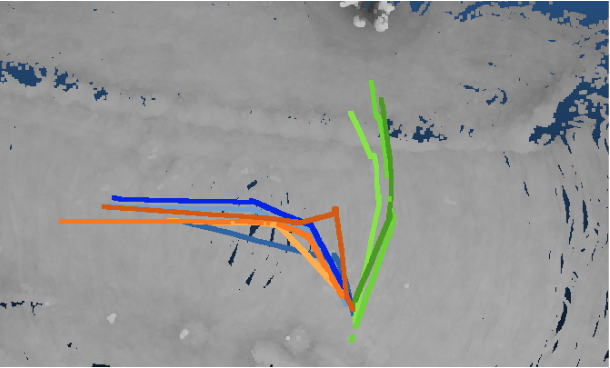}}\;\raisebox{-0.5\height}{\includegraphics[height=1.9cm,width=.19\linewidth]{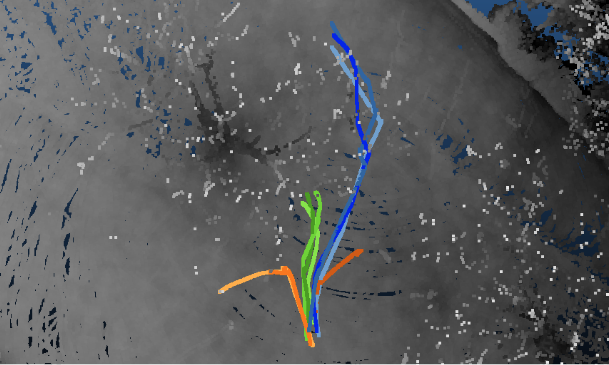}}\;\raisebox{-0.5\height}{\includegraphics[height=1.9cm,width=.19\linewidth]{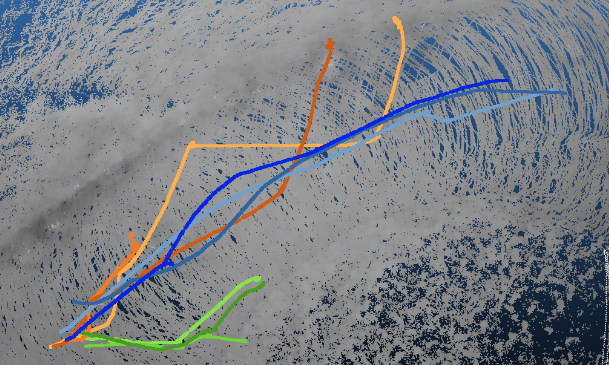}}
    \\\vspace{.2\baselineskip} {\footnotesize (1)} \hspace{3cm}{\footnotesize (2)} \hspace{3cm}{\footnotesize (3)} \hspace{3cm}{\footnotesize (4)} \hspace{3cm}{\footnotesize (5)}
    \vspace{.1cm}
      \caption{Three approaches are tested in five scenes including asphalt roads, grass, curbs, tactile pavings, gravels (top row). The wheelchair starts at the same initial pose and tend to reach the desired goal points (red cross). The bottom row shows recorded wheelchair trajectories in a top-down view (our RCA in blue; semantic-based in red; and 3D-based in green). In (1), RCA detours away from the curb and uneven surfaces until it reaches the ramp. In (2), RCA takes a side way while avoiding the turbulence brought by tactile pavings. In (3)\&(4), RCA foresees the effect from sinking in the soft terrain and avoids the vegetation although going through it is the shortest path. In (5), RCA keeps the wheelchair away from the vegetation and gravels. }%
      \label{fig:scenes}%
      \vspace{-15pt}%
  \end{figure*}
  
\vspace{0.1cm}
\subsubsection{Human Evaluation}\label{subsubsec:human-eval}

    We designed a Amazon Mechanical Turk (AMT) study to assess human evaluation of the performances on five dimensions of ride comfort. 
    During a session, workers are expected to watch three videos showing different robot behaviors in the same scenario; the order of three videos is randomized.
    The participants are asked to rank the three videos from the most preferred (1) to the least preferred (3) according to the following five criteria:
    \begin{itemize}
        \item Stability: 
        A ride is considered stable when noticing 
        few frame-to-frame jitters and  gentle orientation changes.
        \item Path normality: 
        A ride is considered normal if 
        the path matches with a human's expectation.
        \item Safety: 
        A ride is considered as safe when observing continuously smooth motions, predictable paths, and avoiding obstacles at proper distance.
        \item  Trustworthiness: 
        A ride is  considered as trustworthy if a passenger could rely on the wheelchair to complete trips independently.
        \item  Overall preference: 
        A ride is preferred when you would like to have a same wheelchair in your community.
    \end{itemize}



\section{Results}\label{sec:results}

We used 45 robot videos for 5 scenarios shown in~\Cref{fig:scenes}
for each of the three approaches with 3 repeated trials. 
After briefing on the learning performance, we report the PMI results to provide quantitative analysis on our robot experiments. We then report on the full human evaluation results based on 55 MTurk workers' responses. 

\subsection{PMI Analysis}
     \begin{figure}[hb!]%
      \centering%
      \vspace{-15pt}
      \includegraphics[width=0.8\linewidth]{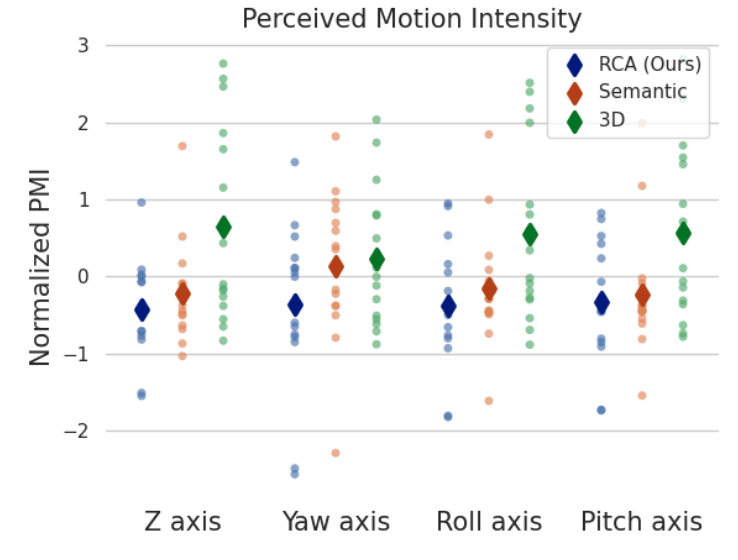}
      \caption{Normalized Perceived Vehicle Motion Intensity Scores\cite{de_winkel_pmi_2020} for Z (upward pointing), yaw, roll and pitch axes. Our approach (blue) shows consistently smaller scores for all axes, demonstrating less disturbances. Semantic-based baseline (orange) exhibits a higher value along Yaw axis, implying its frequent and aggressive turning motions. 3D-based baseline (green) tends to have larger values along Z, Roll, and Pitch axes, indicating its frequent shakiness over uneven terrains.}%
      \label{fig:pmi}%
  \end{figure}
  
\Cref{fig:pmi} shows the aggregated PMI (\Cref{eq:pmi}) results from averaging each run of the robot experiments. Our approach brought less turbulence and unstable motions to the vehicle, especially in the vertical, roll, and pitch axes. As traversal cost is defined based on the amount of variations in the vehicle states, it inherently switches to regions with less intense displacements along three axes. Limited by the angle of view from the front-facing camera, the wheelchair displays occasional turnings along long trips. Although it introduces a fair amount of yaw axis changes to avoid uneven terrains, the PMI originated from heading changes is relatively comparable as the 3D-based approach, which tends to generate a straight-line path wherever terrain heights permit. The semantic-based method avoids moderately rough terrains; however, it fails to provide fine-grid estimation within a semantic class, and thus results in more frequent turnings to search for alternative passable semantic classes.
      \begin{figure}[b!]%
      \centering%
        \vspace{-15pt}%
      \includegraphics[trim={6cm 0 0 0},clip, width=\linewidth]{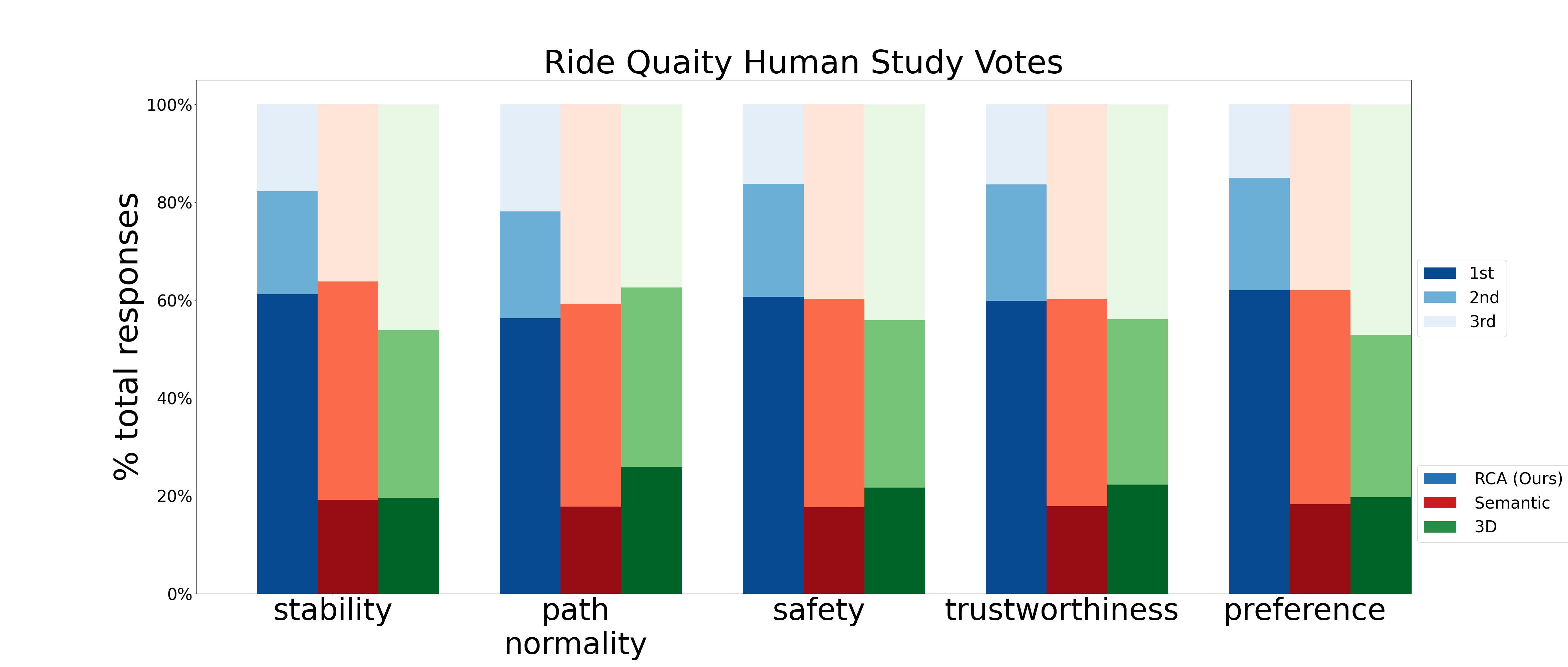}%
      \caption{Percentage of samples ranking three approaches among five dimensions. Each vertical bar consists of three segments representing the percentage of samples voting an approach as the first (bottom, darker color), the second (middle, moderate color), the third (top, lighter color). Our approach marks as the blue bar, while the semantic-based baseline is in red and the 3D-based baseline is in green. Across five evaluation criteria over five scenes, our approach is most likely to be ranked as the top. We fail to show a clear winner between semantic-based approach and 3D-based approach.}%
      \label{fig:votes}%
  \end{figure}
  
\subsection{Human Evaluation Analysis}
    \Cref{fig:votes} presents a vote breakdown in percentage received by each approach among all participants. Based on these collective responses, we found that our approach is substantially better than two baseline approaches on average. The results show consistent top rankings across five dimensions with moderate variations across scenarios. It does not show a clear first runner up approach. The results also cast light on the positive correlation between objective factors (i.e. stability, safety) to subjective factors (i.e. trustworthy, preference).

\subsection{PMI vs. Human Evaluation}\label{subsec:pmi-human}
    Tab.~\ref{tab:spearman} reports the Spearman Correlation scores between stability rankings and Perceived Motion Intensity ranks along Z, Roll, Pitch, Yaw axes among all experiment runs with $p < 0.01$. It reveals strong positive correlations between human perceived stability scores and Perceived Motion Intensity computed from acceleration and jerk along Z, Roll, Pitch, Yaw axes. It further validates that our approach improves the perceived stability relatively by reducing motion intensity along Z, Roll, Pitch, Yaw axes. 
    \begin{table}[h!]
    \begin{center}
    \caption{Spearman Correlation Scores}\label{tab:spearman}
    \begin{tabular}{c|c|c|c|c}
    \hline
      & PMI Z & PMI Roll &  PMI Pitch & PMI Yaw \\
    \hline
    Stability & 0.69 & 0.70 & 0.61 & 0.80 \\
    \hline
    \end{tabular}
    \end{center}
    \end{table}

\section{CONCLUSION}

    In this paper, we introduce a self-supervised ride comfort-aware  approach for terrain traversability analysis. It is motivated by curtailing disturbances caused by traversing uneven or soft surfaces to enhance wheelchair ride comfort. Our RCA approach uses internal vehicle states to self-supervise the process of traversal cost estimation from visual images. According to Perceived Motion Intensity Scores, the results of robot experiments strongly support the effectiveness of our approach on reducing disturbances. Based on human evaluation of 55 MTurk workers, RCA is ranked the highest when compared to the baseline approaches in terms of stability, path normality, safety, trustworthiness, and overall preference. We also recognize that RCA does not perform consistently well across all scenarios. Whereas RCA generally performs well by estimating the disturbing motions brought by uneven or soft terrains, the semantic-based approach provides human-interpretive estimation in the pixel space and the 3D-based approach provides robust and accurate geometric measurements of rigid objects. These findings lead to our future direction for unifying comfort-aware, semantic-based, and 3D-based approaches towards the goal of supporting safe and comfortable rides for the wheelchair users. 




\section*{ACKNOWLEDGMENT}

The authors thank Zhanxin Wu for sharing her implementation of the semantic baseline approach, and Peter Schaldenbrand, Tanmay Shankar, Hyeonwoo Yu for contributing ideas for our design of the framework and the human study. Special thanks to Ziyan Zhang and Zeyu Wang for assisting robot experiments.

\section*{APPENDIX}
We also include additional information for reference.
 \subsection{Dataset Examples}
\begin{figure}[thb]
      \centering
        \vspace{-10pt}
        \raisebox{-0.5\height}{\includegraphics[ width=.3\linewidth]{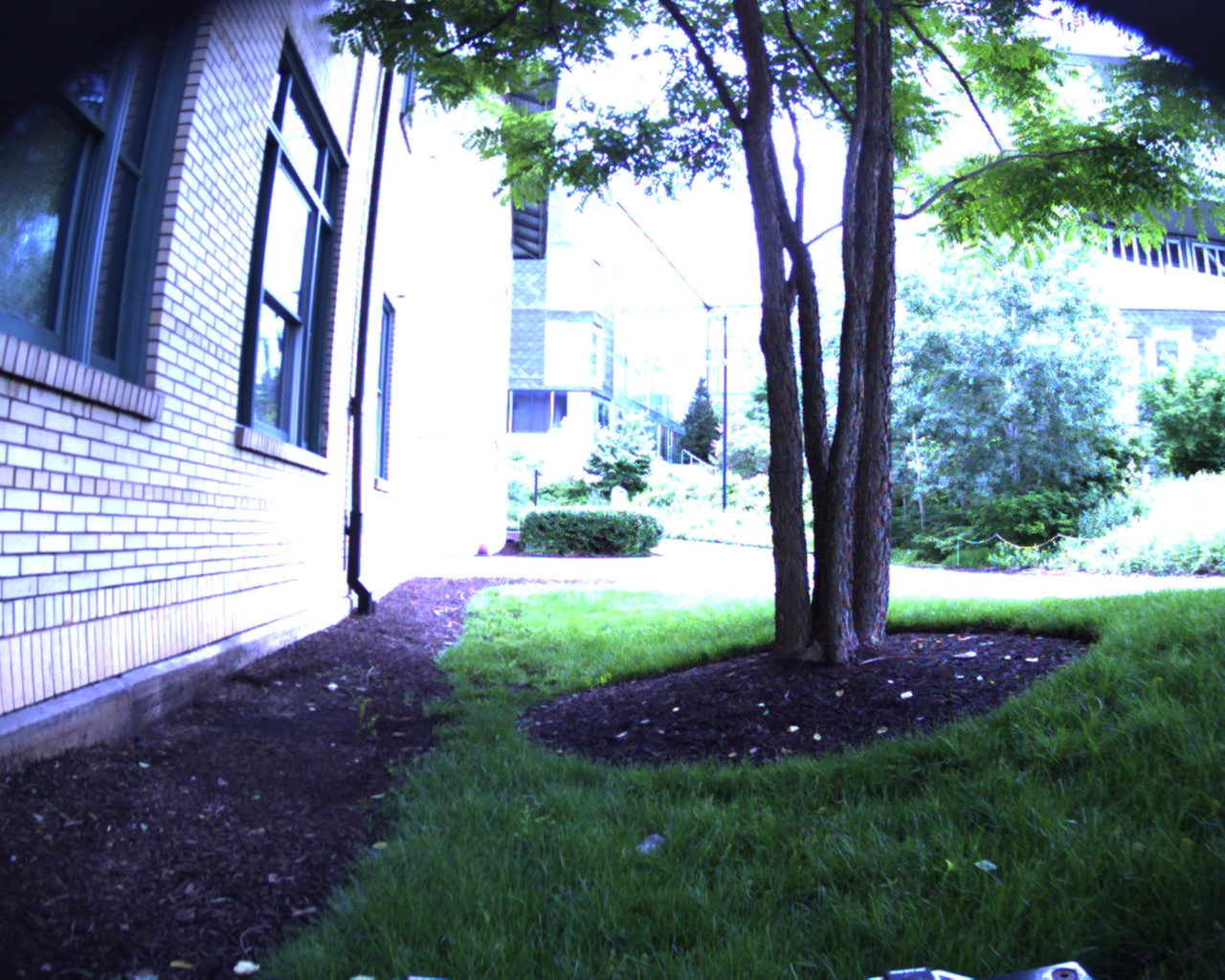}}\;\raisebox{-0.5\height}{\includegraphics[ width=.3\linewidth]{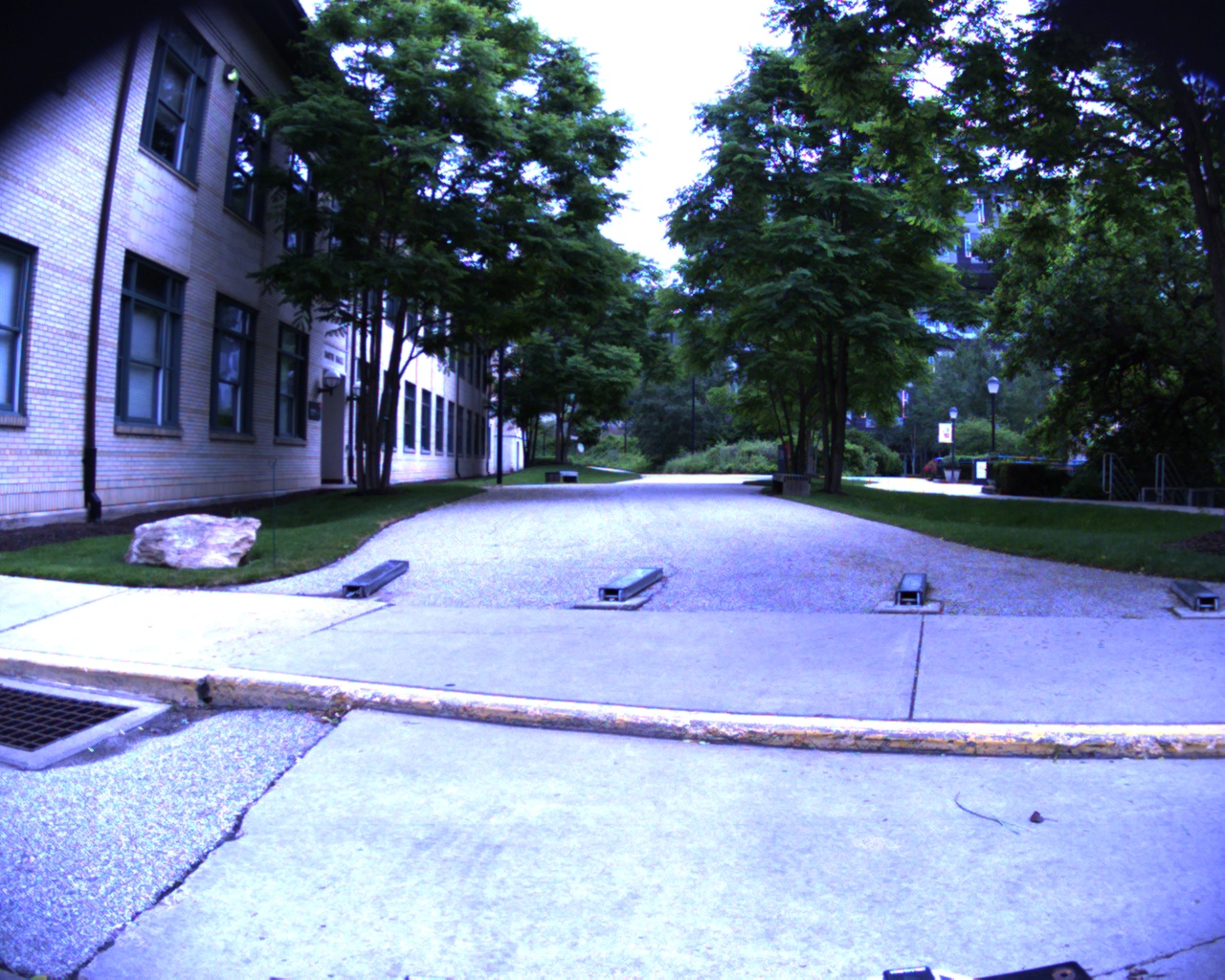}}\;\raisebox{-0.5\height}{\includegraphics[width=.3\linewidth]{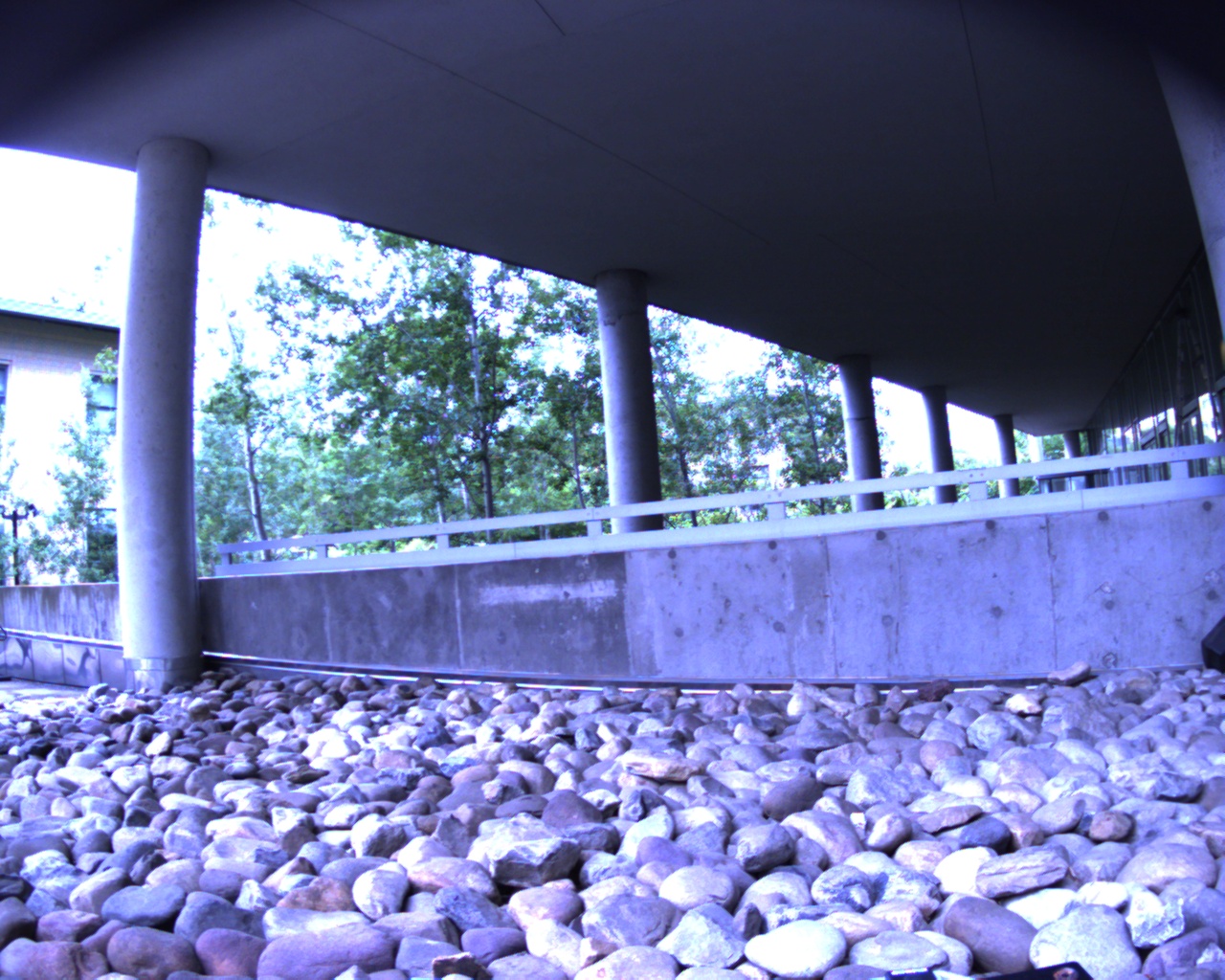}}\;
        \\ \vspace{.2\baselineskip}
        \raisebox{-0.5\height}{\includegraphics[ width=.3\linewidth]{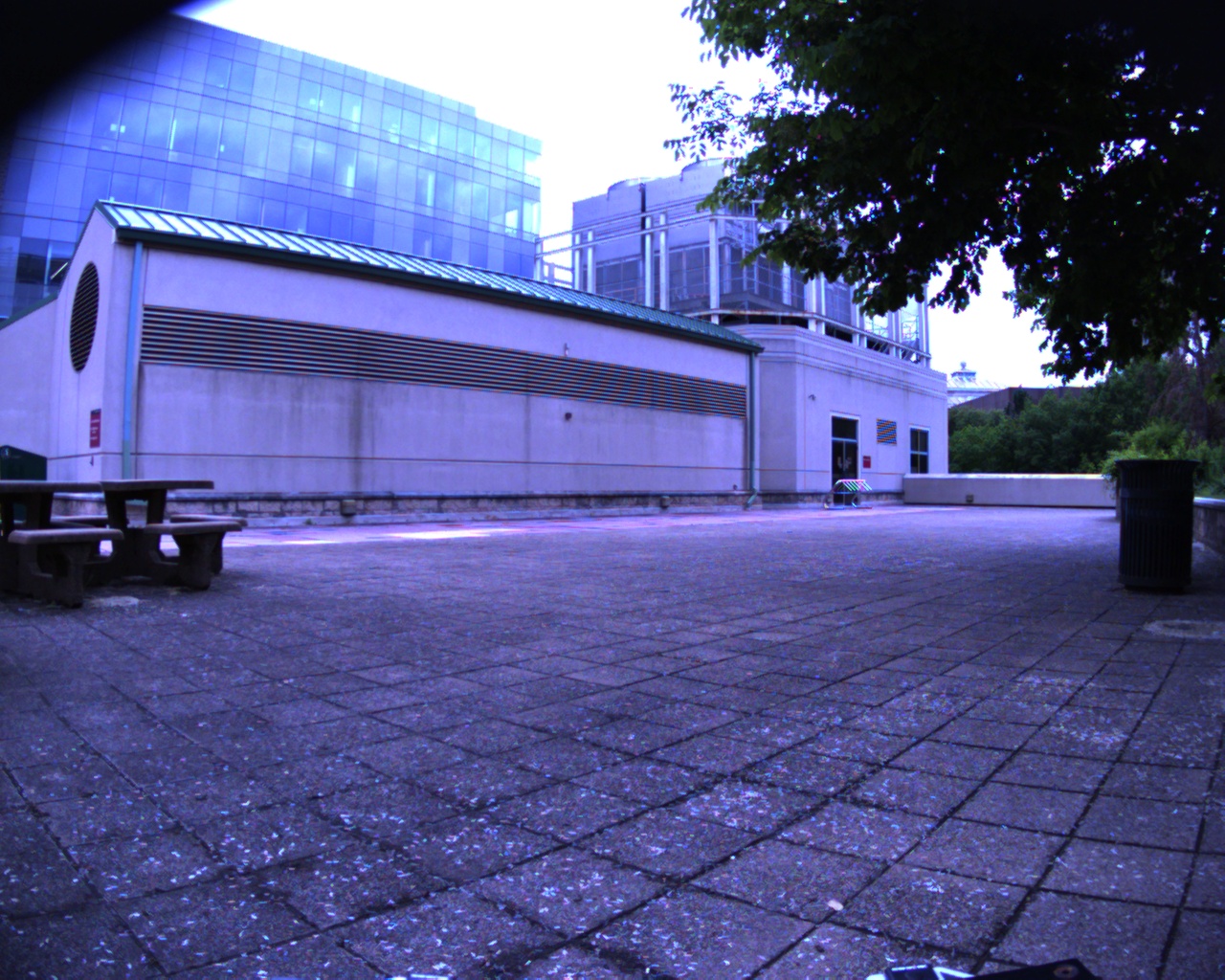}}\;\raisebox{-0.5\height}{\includegraphics[width=.3\linewidth]{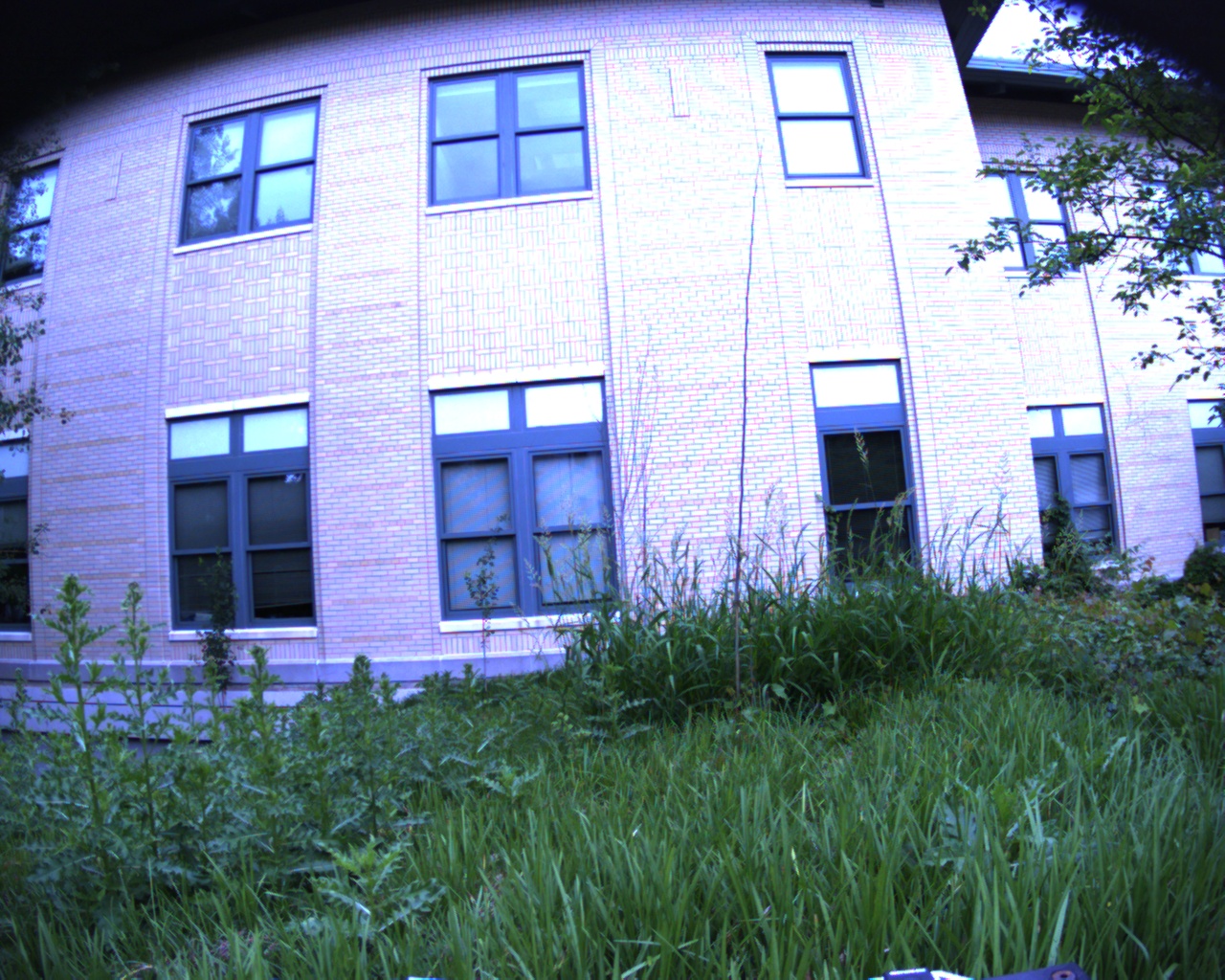}}\;\raisebox{-0.5\height}{\includegraphics[width=.3\linewidth]{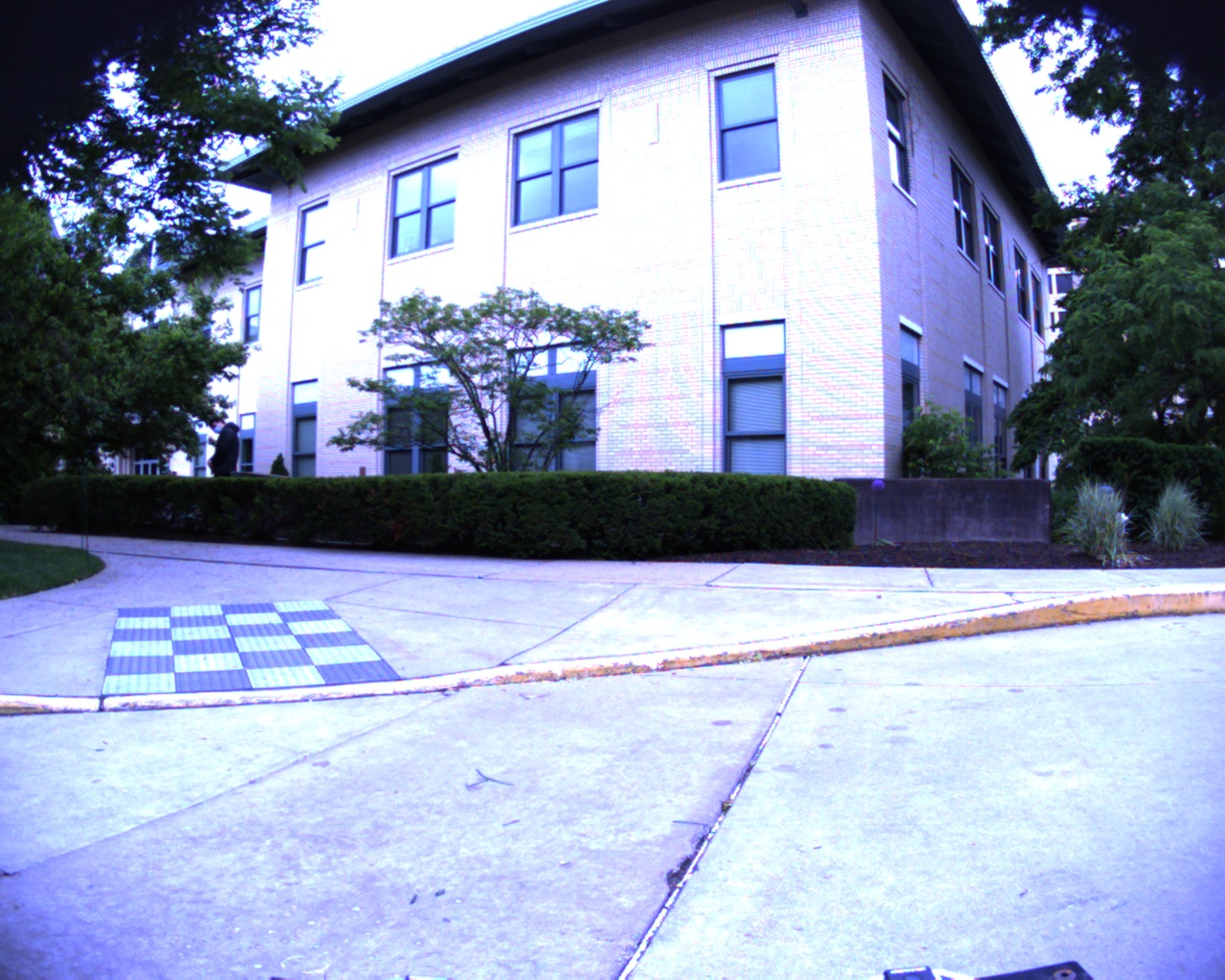}}\;
      \caption{Our dataset includes various types of soft and hard terrains.}%
      \label{fig:ds}%
  \end{figure}

Example scenes of our dataset is shown in \Cref{fig:ds}.
 
 \subsection{Evaluation Metrics for Learning Accuracy}
  
  Following previous works \cite{Zurn2021SelfSupervisedVT}\cite{kris2021recog}, we use two metrics to evaluate classification of vehicle state unsupervised learning using weakly supervised labels. One is normalized mutual information (NMI) defined as:

     \begin{equation*}
         \operatorname{NMI}(Y, C)=\frac{2 I(Y, C)}{H(Y)+H(C)}  
     \end{equation*}
 where $I(Y, C)$ denotes the mutual information between cluster sets $Y$ and $C$, and $H(Y)$, $H(C)$ denote the entropy of cluster sets $Y$ and $C$ respectively. We define $Y$ as the cluster set of the ground truth labels. The clustering accuracy can be defined as:
     \begin{equation*}
     \operatorname{Accuracy}(Y, C)=\frac{1}{N} \sum_{k} \max _{j}\left|y_{k} \cap c_{j}\right|
     \end{equation*} 

For evaluating self-supervised learning for cost prediction, we also measure how well vehicle states are reconstructed given the visual input to indicate how much the feature domain is translated from image to vehicle states. We use Mean Square Error (MSE) average over three axes defined as:
\begin{equation*}
\mathrm{MSE}=\frac{1}{3}\sum_{d=1}^{3}\left(\frac{1}{n} \sum_{i=1}^{n}\left(\mathbf{A^d_{i}}-\mathbf{\hat{A}^d_{i}}\right)^{2}\right)
 \end{equation*}  

 \subsection{Learning Verification Results}  
\begin{figure}[thb]%
      \centering%
        \vspace{-10pt}
      \includegraphics[width=0.7\linewidth]{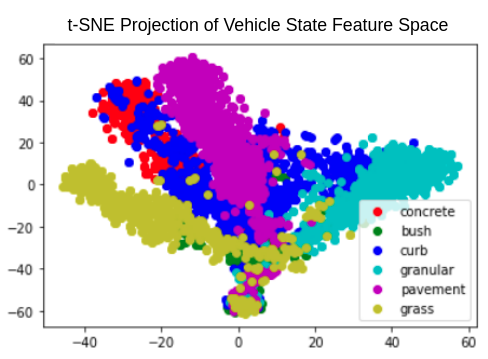}
      \caption{Visualization of vehicle state feature space using t-SNE projection. Samples are colored by semantic classes of recorded scenes.}%
      \label{fig:tsne}%
  \end{figure}

For our verification purpose, we first visualize the vehicle state feature space colored by semantic classes of recording scenes (\Cref{fig:tsne}). We note that during the unsupervised learning process, those semantics classes serve as a prior for triplet selection. The visualization validates our hypothesis that semantic classes do not fully represent the inherent structure of vehicle dynamics features, and thus our RCA approach based on vehicle states is more targeted toward safe and comfortable navigation.

\begin{figure*}[thb]%
    \centering
\raisebox{-0.5\height}{\includegraphics[width=0.45\linewidth]{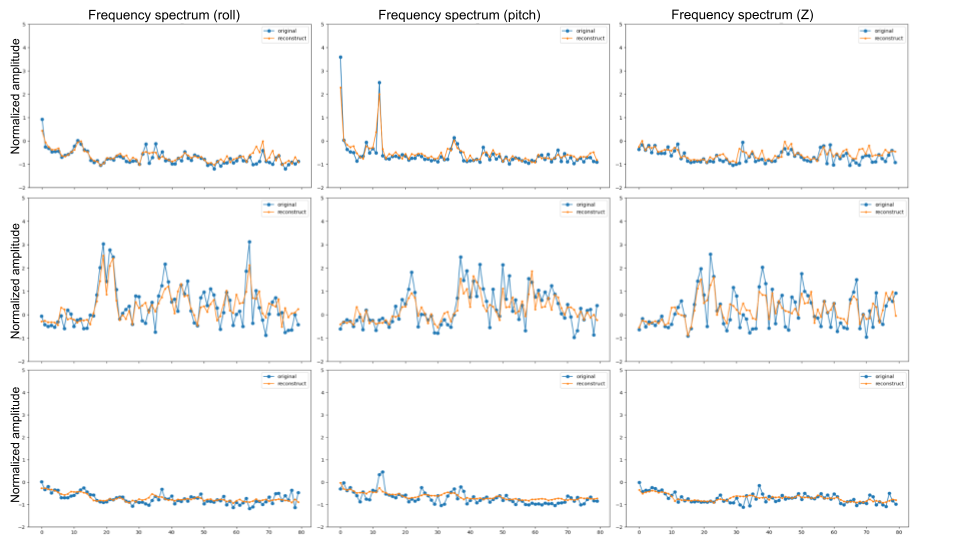}}\;\raisebox{-0.5\height}{\includegraphics[width=0.45\linewidth]{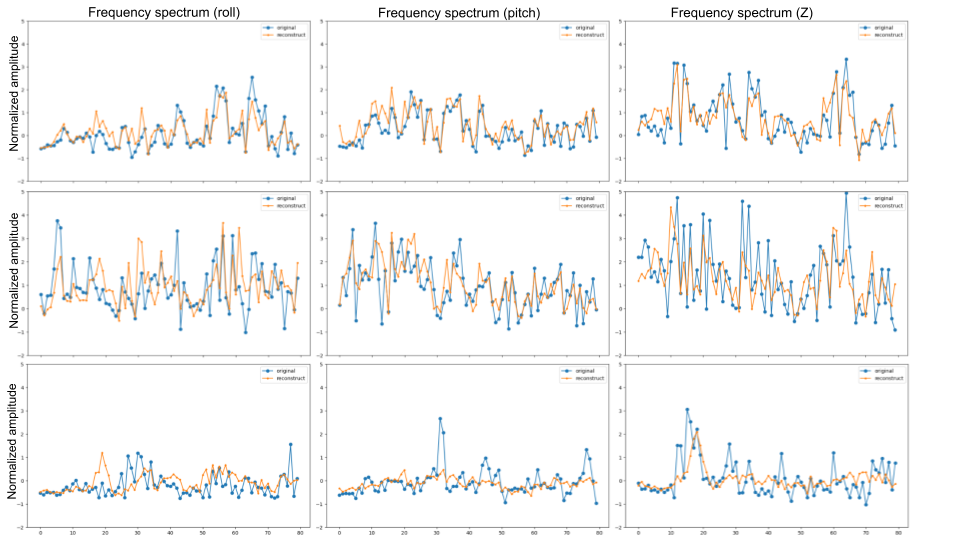}}
    \caption{Prediction results of normalized amplitude spectra along roll, pitch and Z axis. It shows relative amounts of amplitude for each frequency component. Ground truth signals are marked in blue, while the estimates ones are shown in orange.}
    \vspace{-20pt}
    \label{fig:mse}
    \vspace{10pt}
    \end{figure*}

We then measure the performance of unsupervised vehicle state clustering in terms of NMI and Accuracy, $0.56$ and $86.84$, respectively. Next, the performance of the vehicle state prediction, measured by MSE, was $0.53$. \Cref{fig:mse} qualitatively shows the prediction results against the ground truth. The main evaluation was done through the downstream task of robot navigation. 

\bibliographystyle{./IEEEtran}
\bibliography{./IEEEabrv,./IEEEexample}

\end{document}